\documentclass[10pt,twocolumn,letterpaper]{article}

\usepackage{cvpr}
\usepackage{times}
\usepackage{epsfig}
\usepackage{graphicx}
\usepackage{amsmath}
\usepackage{amssymb}
\usepackage{floatrow}
\usepackage{gensymb}

\newfloatcommand{capbtabbox}{table}[][\FBwidth]

\usepackage[pagebackref=true,breaklinks=true,letterpaper=true,colorlinks,bookmarks=false]{hyperref}

\cvprfinalcopy 

\def\cvprPaperID{****} 

\ifcvprfinal\fi
\begin{document}

\title{Lightweight Real-time Makeup Try-on in Mobile Browsers with Tiny CNN Models for Facial Tracking}

\author{TianXing Li
\thanks{These two authors contributed equally.}
\thanks{The work performed during internship at ModiFace Inc.}\\
ModiFace Inc.\\
{\tt\small tianxingli24@gmail.com}
\and
Zhi Yu \footnotemark[1]\\
ModiFace Inc.\\
{\tt\small vicky@modiface.com}
\and
Edmund Phung\\
ModiFace Inc.\\
{\tt\small edmund@modiface.com}
\and
Brendan Duke\\
ModiFace Inc.\\
{\tt\small brendan@modiface.com}
\and
Irina Kezele\\
ModiFace Inc.\\
{\tt\small irina@modiface.com}
\and
Parham Aarabi\\
ModiFace Inc.\\
{\tt\small parham@modiface.com}
}

\maketitle

\begin{abstract}
  Recent works on convolutional neural networks (CNNs) for facial alignment have demonstrated unprecedented accuracy on a variety of large, publicly available datasets.
  However, the developed models are often both cumbersome and computationally expensive, and are not adapted to applications on resource restricted devices. In this work, we look into developing and training compact facial alignment models that feature fast inference speed and small deployment size, making them suitable for applications on the aforementioned category of devices.
  Our main contribution lies in designing such small models while maintaining high accuracy of facial alignment.
  The models we propose make use of light CNN architectures adapted to the facial alignment problem for accurate two-stage prediction of facial landmark coordinates from low-resolution output heatmaps.
  We further combine the developed facial tracker with a rendering method, and build a real-time makeup try-on demo that runs client-side in smartphone Web browsers.
  We prepared a demo link to our Web demo that can be tested in Chrome and Firefox on Android, or in Safari on iOS: \url{https://s3.amazonaws.com/makeup-paper-demo/index.html}\footnote{The link works in listed browsers on corresponding devices, and should start with ``https://''.}\footnote{The project website is: \url{http://research.modiface.com/makeup-try-on-cvprw2019/}}
\end{abstract} 
\vspace{-5mm}
\section{Introduction}
\vspace{-2mm}
Facial alignment is a key component of virtual makeup try-on methods. These methods require very high accuracy in facial alignment, as any error in makeup placement would result in a poor user experience. The alignment also needs to be robust to variations in lighting, pose, face shape, and skin tone. It is particularly important that the landmarks are precisely aligned for frontal and close to frontal face poses, as those are common in virtual try-on applications. \\
\indent In the context of real-time applications on resource constrained platforms such as mobile and Web, we need to address the requirements for both low computational demands and small model size. 
The latter is of particular importance for client-side Web applications where long loading times are not ideal.\\
\indent The state-of-the-art facial alignment architectures \cite{deepalignmentnet}\cite{deepconvnetcascade}\cite{stackedhourglass} have not been developed for the purpose of real-time inference on resource restricted devices. However, our applications primarily target such devices. To strike a better balance for real-time applications on those platforms, the ideal architecture should minimize load and execution time while preserving alignment accuracy. \\
\indent Our proposed architecture does the following to meet these requirements:
its first stage makes coarse initial predictions, from which crops of shared convolutional features are taken; these Regions of Interest (RoIs) are then processed by the second stage to produce more spatially refined predictions. 
Besides the coarse-to-fine alignment approach, we also balance the number of layers and the resolution of feature maps,
and achieve fine-level alignment with high computational efficiency, while maintaining a small model size.
The resulting architecture is suitable for real-time Web applications in mobile browsers.
\vspace{-1mm}
\section{Related Work}
\label{sec:prior}
\vspace{-1mm}
\subsection{Facial Landmark Alignment}
\vspace{-2mm}
\indent The facial landmark alignment problem has a long history with classical computer vision solutions. 
For instance, the fast ensemble tree based \cite{Kazemi2014OneMF} algorithm achieves reasonable accuracy and is widely used for real-time face tracking \cite{dlib09}. 
However, the model size required to achieve such accuracy is prohibitively large.\\
\indent Current state-of-the-art accuracy for facial landmark alignment are achieved by convolutional neural network based methods. 
To maximize accuracy on challenging datasets \cite{LFPW}\cite{helen}\cite{ibug}, they use large neural networks that are not real-time, and have model sizes of tens to hundreds of MB \cite{stackedhourglass}\cite{hourglass} that would entail unreasonable load times for Web applications. 
\vspace{-2mm}
\subsection{Efficient CNN Architectures}
\vspace{-2mm}
\indent To bring the performance of convolutional neural networks to mobile vision applications, numerous architectures with efficient building blocks such as MobileNetV2 \cite{mobilenetv2}, SqueezeNet \cite{squeezenet} and ShuffleNet \cite{shufflenet} have recently been released. 
These networks aim to maximize performance (e.g., classification accuracy) for a given computational budget, which consists of the number of required learnable parameters (the model size) and multiply-adds. \\
\indent We will focus our discussion on the state-of-the-art MobileNetV2, whose inverted residual blocks are used in our own design. 
Their choice of depthwise convolutions over regular convolutions drastically reduces the number of multiply-adds and learnable parameters, at a slight cost in performance \cite{mobilenetv2}.
Furthermore, the inverted design, which is based upon the principle that network expressiveness can be separated from capacity, allows for a large reduction in the number of cross-channel computations within the network \cite{mobilenetv2}.\\
\indent Finally, the residual design similar to ResNet \cite{resnet} eases issues with gradient propagation in deeper networks. 
\vspace{-2mm}
\subsection{Heatmap}
\vspace{-2mm}
\indent Fully convolutional neural network architectures based on heatmap regression \cite{humanPoseHeatMap}\cite{convposemachine}\cite{chen2019adversarial}\cite{pishchulin2016deepcut} have been widely used on human pose estimation tasks. The use of heatmaps provides a high degree of accuracy, along with an intuitive means of seeing the network's understanding and confidence of landmark regression. 
This technique has also been used in recent facial alignment algorithms such as the Stacked Hourglass architecture \cite{stackedhourglass}. 
However, the Stacked Hourglass approach \cite{stackedhourglass} uses high resolution heatmaps, which require a large amount of computation in the decoding layers. 
There is room for optimization here, as the heatmaps only have non-negligible values in a very concentrated and small portion of the overall image. 
This observation motivates us to use regional processing, which allows for the network to focus its processing in relevant areas. (i.e., the approximate Rregion of Interest).
\vspace{-2mm}
\subsection{Mask RCNN}
\vspace{-2mm}
\indent There are a series of frameworks which are flexible and robust to object detection and semantic segmentation like Fast R-CNN \cite{girshick2015fast}, Faster R-CNN \cite{ren2015faster} and Fully Convolutional Network \cite{long2015fully}. 
Faster R-CNN proposes a multi-branch design to perform bounding box regression and classification in parallel. 
Mask-RCNN \cite{maskrcnn} is an extension of Faster-RCNN, and adds a new branch for predicting segmentation masks based on each Region of Interest. 
Of particular interest is Mask-RCNN's use of RoI align \cite{maskrcnn}, which allows for significant savings in computation time by taking crops from shared convolutional features. 
By doing this, it avoids recomputing features for overlapping Regions of Interest.
\vspace{-2mm}
\section{Proposed Method}
\vspace{-2mm}
\indent There are two main contributions of our model:
\vspace{-2mm}
\begin{enumerate}
\setlength\itemsep{0em}
\item We use RoI align \cite{maskrcnn} for each individual landmark to save potentially overlapping computation, allow the network to avoid non-relevant regions, and force the network to learn to produce useful shared features.
\vspace{-1mm}
\item Our two-stage localization architecture along with auxiliary coordinate regression loss allow us to work with extremely small and computationally cheap heatmaps at both stages.
\end{enumerate}
\vspace{-3mm}
\subsection{Model Structure}
\begin{figure}[hbt!]
\centering
\includegraphics[width=0.9\textwidth]{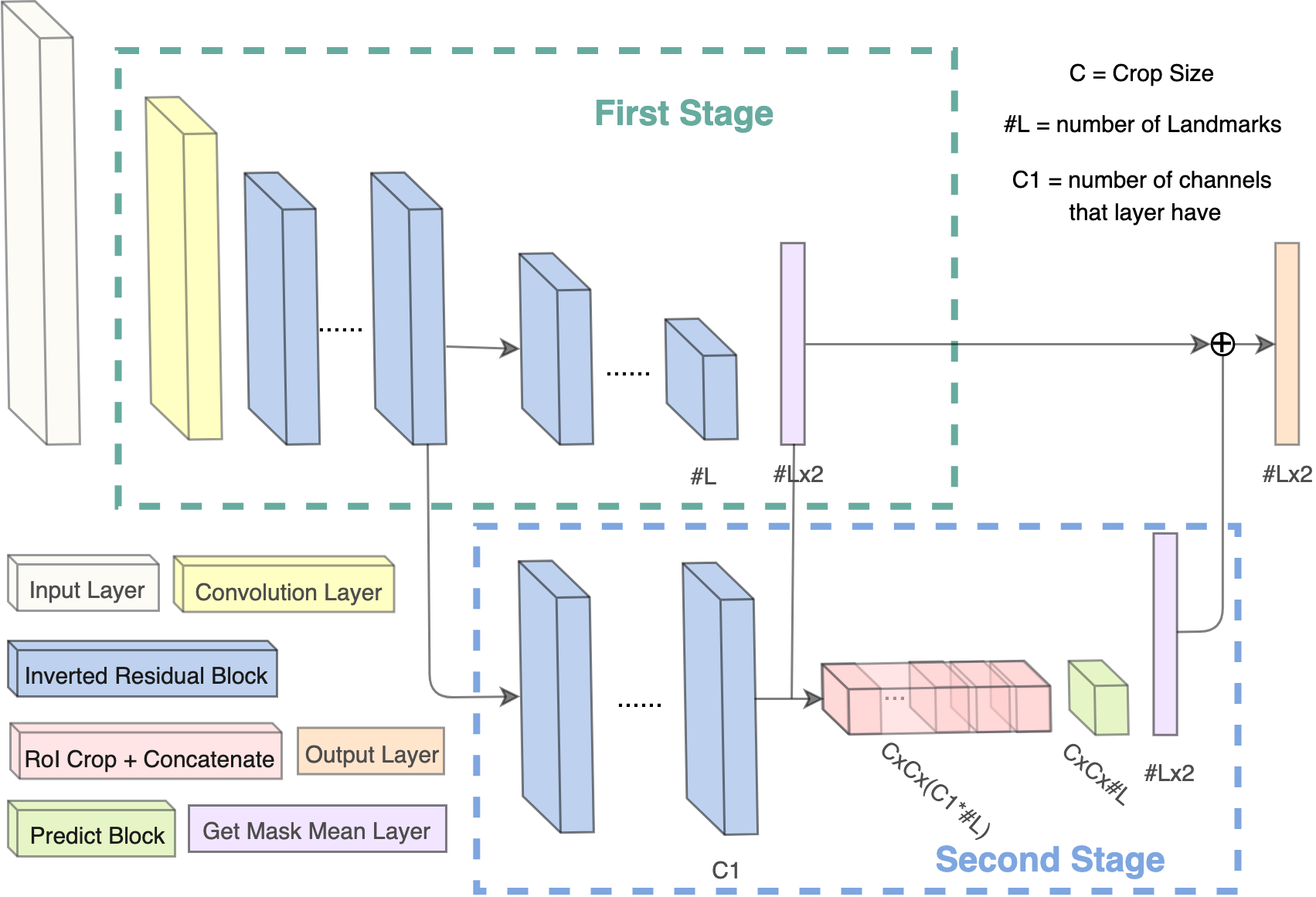}
\vspace{-3mm}
\caption{\label{fig:network}Our two-stages Network Architecture.}
\end{figure}

\vspace{-2mm}
\indent The model has two-stages and is trained end-to-end, as illustrated in Figure \ref{fig:network}.
\vspace{-3mm}
\vspace{-4mm}
\\
\indent The first stage is formed by a list of Inverted Residual Blocks \cite{mobilenetv2}, which predict $H \times H$ heatmaps, one for each facial landmark. Interpreting the normalized activations over the heatmaps as a probability distribution, we compute the expected values of these heatmaps to obtain the $x, y$ coordinates. 
This will be described in more detail in the following subsection.\\
\indent The second stage has several shared convolutional layers, which branch off from part of the first stage.
Using the coarse predictions from the previous stage, we apply RoI align \cite{maskrcnn} to the final shared convolutional features. 
Each of the cropped features are input to one final convolutional layer, which has separate weights for each individual landmark. 
Our predict block implements this efficiently with the use of group convolutions \cite{shufflenet}. 
The final output is a heatmap for each landmark. The coordinates obtained from these heatmaps indicate the required offset from the initial coarse prediction, i.e., if the heatmap at this stage is perfectly centered, then there is effectively no refinement applied to the first stage prediction. 
\vspace{-2mm}
\subsection{Coordinate Regression from Heatmaps}
\vspace{-3mm}
\indent For our ground truth heatmaps, we use a Gaussian distribution with a mode corresponding to the ground truth coordinates' positions. 
Letting $x, y$ denote the coordinates of any pixel in the feature map, we can compute its value using a $2D$ Gaussian distribution with the corresponding landmark coordinate as center.\\
\indent The regressed $x_{pred}, y_{pred}$ is then the expected value of the pixel locations according to the distribution computed from the heatmap. Let $i,j$ denote the coordinates in the heatmap, and $\rho_{ij}$ denote the corresponding probability value:\\
\vspace{-5mm}
\begin{equation}
\small
\begin{bmatrix}
x_{pred}, 
y_{pred} \\
\end{bmatrix} = \sum^{W}_{i=1}\sum^{H}_{j=1}\rho_{ij}
\begin{bmatrix}
x_{i},
y_{j} \\
\end{bmatrix}
\end{equation}
\vspace{-6mm}

\subsection{Loss Function}
\vspace{-3mm}
\indent We apply a pixelwise sigmoid cross entropy \cite{pixelwisesigmoidcrossentropy} to learn the heatmaps, which we denote as $L_h$. 
Additionally, in order to alleviate issues with the heatmaps being cut off for
landmarks near boundaries, we add on an $L_2$ distance loss with a loss weight~$\lambda$: $L_{total} = L_h + \lambda\cdot L_2$.
\vspace{-2mm}
\begin{equation}\label{eq:heatmap_loss}
\small
\begin{split}
L_h =&\space \frac{1}{N}\sum^{N}_{n=1}\sum^{L}_{l=1}\sum^{W}_{i=1}\sum^{H}_{j=1}\\
    & [\rho_{ij}^{l}log\hat{\rho^{l}_{ij}} + (1 - \rho_{ij}^{l})log(1 - \hat{\rho^{l}_{ij}})]\cdot w_{ij}^l 
\end{split}
\end{equation}
\vspace{-2mm}
\begin{equation}\label{eq:weight}
\small
w_{ij}^{l} = \frac{((i^n-\hat{i_l^n})^2 + (j^n-\hat{j_l^n})^2)\cdot 2}{W^2+H^2}
\end{equation}
Where $\rho_{ij}^{l}$ is the prediction value of the heatmap in the $l$th channel at pixel location $(i,j)$ of $n$'s sample, while $\hat{\rho^{l}_{ij}}$ is the corresponding ground truth. $w_{ij}^{l}$ is the weight at that location, which is calculated from Equation~\ref{eq:weight}.
$(\hat{i_l^n}, \hat{j_l^n})$ is the ground truth coordinate of the $n$'s sample's $l$th landmark.
\vspace{-2mm}
\subsection{Data}
\vspace{-3mm}
\indent Our dataset consists of a union of Helen \cite{helen}, LFPW \cite{LFPW} and iBUG \cite{ibug} datasets and additional images that we collected in-house, with $3681$ images in total. The image annotation follows Helen annotation format \cite{helen}, but with corrected original annotations and with an addition of a couple of new eyebrow landmarks. The contour annotation is sparse. This dataset has a total of 62 inner points and 3 contour points as shown in Figure \ref{fig:annotaion}. The pose distribution is approximately -30\degree to 30\degree  in yaw angle, which agrees well with our application scenario. 
For the purpose of comparison with state-of-the-art methods, we further train and test our model design on the 300W dataset \cite{sagonas2013300}.
\vspace{-2mm}
\section{Results and Ablation Study}
\begin{figure}[hbt!]
\begin{floatrow}
\ffigbox{%
  \includegraphics[width=0.3\textwidth]{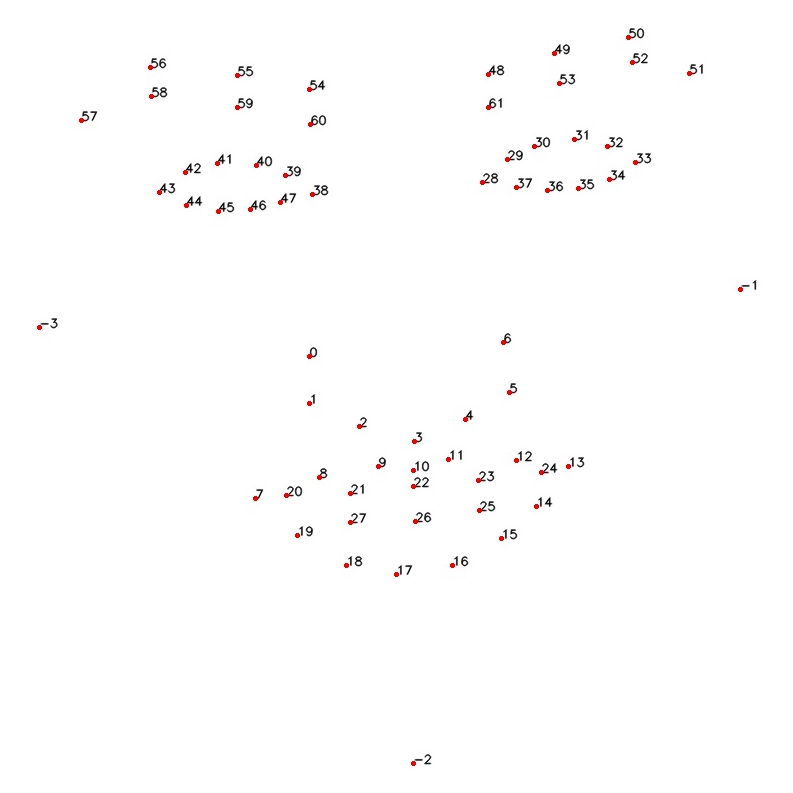}
}{%
  \vspace{-2mm}
  \caption{\label{fig:annotaion}The annotation of~65 landmarks.}%
}
\capbtabbox{%
\begin{tabular}{cc} \hline
\hline
Total params& 158,091\\
\hline
Total MAdd& 173.69M\\
\hline
Total Flops& 90.75M\\
\hline
Model Size& 607KB\\
\hline
\begin{tabular}[c]{@{}l@{}}Inference Time \\ (iPhone XR)\end{tabular} & 20ms\\
\end{tabular}
}{%
  \vspace{-2mm}
  \caption{Our demo model, results in Table \ref{tab:table1} last row.}
\label{tab:table2}%
}
\end{floatrow}
\end{figure}

\vspace{-3mm}
The reported error is the commonly used mean squared error normalized by the inter-pupil distance. 
Table \ref{tab:table1} shows the full model error (of the two-stage model trained with the heatmap loss, combined with $L_2$ loss) in the last row, and the results when only $L_2$ loss (first row), or only heatmap loss (second row), or only one CNN stage (third row) are used. This ablation study clearly shows that all three tested components: heatmap loss, $L_2$ loss, and extending the architecture with a second processing stage, bear individual importance and improves the model accuracy. \\
\indent The tests on 300W dataset and comparisons with Look at Boundary (LAB) \cite{wu2018look}, a state-of-the-art method for accurate facial alignment, resulted in the following (Table \ref{tab:table3}): our full-size model of 6.6MB achieves similar accuracy as LAB method on 300W Common Subset, while exhibiting some drop in accuracy on the 300W Full dataset, which can be explained by our architectural design being adapted more towards landmark detection on unoccluded and mostly frontal faces.
In addition, ``half~$\alpha$'' ($\alpha = 0.5$), which halves the number of channels per layer, preserves the accuracy when tested on the Common Subset.
Finally, we show that a further reduction in model size and input image resolution does not significantly degrade the accuracy, as shown in Table \ref{tab:table2}, while attaining much faster computational speeds and overall a smaller model size.
\vspace{-3mm}
\begin{table}[hbt!]
\begin{center}
\begin{tabular}{|l|c|c|}
\hline
Method & Inner Error & Contour Error\\
\hline\hline
Without heatmap loss & 3.53 & 9.01\\
Without $L_2$ loss & 4.45 & 12.3\\
Without second stage & 3.50 & 9.32\\
Our full model & \textbf{3.21} & \textbf{9.00}\\
\hline
\end{tabular}
\vspace{-5mm}
\small
\caption{Results on our dataset with 65 points.}
\label{tab:table1}
\end{center}
\end{table}
\vspace{-7mm}

\begin{table}[hbt!]
\begin{center}
\begin{tabular}{llll}
\hline
\multicolumn{1}{l|}{Method}               & \begin{tabular}[c]{@{}l@{}}Common \\ Subset\end{tabular} & \begin{tabular}[c]{@{}l@{}}Challenging \\ Subset\end{tabular} & Fullset \\ \hline
\multicolumn{1}{l|}{LAB (4-stack)\cite{wu2018look}}        & 4.20      & 7.41   & 4.92    \\
\multicolumn{1}{l|}{LAB (8-stack)\cite{wu2018look}}    & \textbf{3.42}   & \textbf{6.98}    & \textbf{4.12}    \\ \hline
\multicolumn{1}{l|}{Our model ($\alpha=1$)}   & \textbf{3.42}  & 8.51          & 4.42    \\
\multicolumn{1}{l|}{Our model ($\alpha=0.5$)} & 3.43       & 14.7        & 5.64   \\ \hline
\end{tabular}
\vspace{-6mm}
\caption{Results on 300W dataset.}
\label{tab:table3}
\end{center}
\end{table}
\vspace{-6mm}

\vspace{-8mm}
\section{Rendering}
\begin{figure}[hbt!]
\centering
\includegraphics[width=0.95\textwidth]{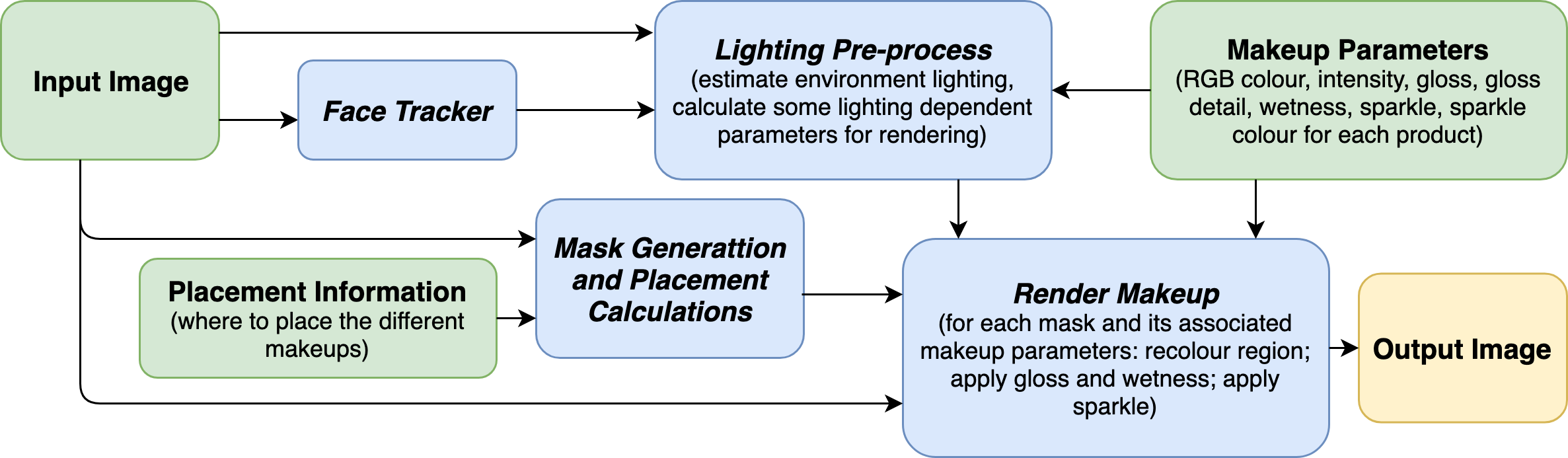}
\vspace{-2mm}
\caption{\label{fig:render}The Rendering Pipeline.}
\end{figure}

\vspace{-3mm}
Following landmark prediction by our CNN model, we define facial parts (e.g., lips) and render makeup on them. 
Figure \ref{fig:render} depicts the full pipeline, included rendering modules. 
\vspace{-3mm}
\vspace{-6mm}
\section{Acknowledge}
We thank Brittany Wanka, Rena Lu and Sarah Koehler for assistance with cleaning and annotating data. We would also like thank Irene Jiang for comments and discussions.
{\small
\bibliographystyle{ieee_fullname}
\bibliography{reference}
}

\end{document}